\documentclass{article}

\usepackage{times}
\usepackage[pdftex]{graphicx}
\usepackage{subfigure} 
\usepackage{natbib}
\usepackage{algorithm}
\usepackage{algorithmic}

\usepackage{amsmath, amsthm}
\usepackage{amsfonts}
\usepackage{amssymb}

\usepackage{hyperref}
\usepackage{dsfont}
\usepackage{overpic}

\usepackage[pdftex]{graphicx} 
\graphicspath{{figures/}} 

\usepackage{algorithm} 
\usepackage{algorithmic}
\usepackage{balance}
\usepackage{url}

\usepackage{soul}

\newcommand{\E}{\mathbb{E}} 
\newcommand{\FE}{\mathcal{F}} 

\DeclareMathOperator{\diag}{diag} 

\newcommand{\expect}[2]{\mathbb{E}_{#1}\left[#2\right]}
\newcommand{\N}{\mathcal{N}} 
\newcommand{\KL}{ {\rm I\!D}_{\text{KL}}}
\newcommand{\Reals}{ {\rm I\! R}}

  \renewenvironment{thebibliography}[1]{%
    \begin{oldthebibliography}{#1}%
      \setlength{\parskip}{0.5ex}%
      \setlength{\itemsep}{0ex}%
}{\end{oldthebibliography}}

\newcommand{\myvec}[1]{\mbox{$\mathbf{#1}$}}
\newcommand{\myvecsym}[1]{\mbox{$\boldsymbol{#1}$}}

\newcommand{\vzero}{\mbox{$\myvecsym{0}$}}

\newcommand{\valpha}{\mbox{$\myvecsym{\alpha}$}}

\newcommand{\vGamma}{\mbox{$\myvecsym{\Gamma}$}}
\newcommand{\vmu}{\mbox{$\myvecsym{\mu}$}}

\newcommand{\vphi}{\mbox{$\myvecsym{\phi}$}}

\newcommand{\vtheta}{\mbox{$\myvecsym{\theta}$}}

\newcommand{\vsigma}{\mbox{$\myvecsym{\sigma}$}}

\newcommand{\vomega}{\mbox{$\myvecsym{\omega}$}}

\newcommand{\vxi}{\mbox{$\myvecsym{\xi}$}}

\newcommand{\vg}{\mbox{$\myvec{g}$}}

\newcommand{\vu}{\mbox{$\myvec{u}$}}

\newcommand{\vw}{\mbox{$\myvec{w}$}}

\newcommand{\vx}{\mbox{$\myvec{x}$}}

\newcommand{\vy}{\mbox{$\myvec{y}$}}

\newcommand{\vz}{\mbox{$\myvec{z}$}}

\newcommand{\vD}{\mbox{$\myvec{D}$}}

\newcommand{\vF}{\mbox{$\myvec{F}$}}
\newcommand{\vG}{\mbox{$\myvec{G}$}}

\newcommand{\vI}{\mbox{$\myvec{I}$}}
\newcommand{\vJ}{\mbox{$\myvec{J}$}}

\newcommand{\vM}{\mbox{$\myvec{M}$}}

\usepackage[accepted]{icml2015}

\icmltitlerunning{Variational Inference with Normalizing Flows}

\begin{document}
\twocolumn[
\icmltitle{Variational Inference with Normalizing Flows}
\icmlauthor{Danilo Jimenez Rezende}{danilor@google.com}
\icmlauthor{Shakir Mohamed}{shakir@google.com}
\icmladdress{Google DeepMind, London}
\icmlkeywords{machine learning, variational inference, normalizing flow}
\vskip 0.3in
]

\begin{abstract} 
The choice of approximate posterior distribution is one of the core problems in variational inference. Most applications of variational inference employ simple families of posterior approximations in order to allow for efficient inference, focusing on mean-field or other simple structured approximations. 
This restriction has a significant impact on the quality of inferences made using variational methods. We introduce a new approach for specifying flexible, arbitrarily complex and scalable approximate posterior distributions. Our approximations are distributions constructed through a normalizing flow, whereby a simple initial density is transformed into a more complex one by applying a sequence of invertible transformations until a desired level of complexity is attained. We use this view of normalizing flows to develop categories of finite and infinitesimal flows and provide a unified view of approaches for constructing rich posterior approximations. We demonstrate that the theoretical advantages of having posteriors that better match the true posterior, combined with the scalability of amortized variational approaches, provides a clear improvement in performance and applicability of variational inference.
\vspace{-3mm}
\end{abstract} 

\vspace{-5mm}
\section{Introduction}
There has been a great deal of renewed interest in variational inference as a means of scaling probabilistic modeling to increasingly  complex problems on increasingly larger data sets. 
Variational inference now lies at the core of large-scale topic models of text \citep{hoffman2013stochastic}, provides the state-of-the-art in semi-supervised classification \citep{kingma2014semi}, drives the models that currently produce the most realistic generative models of images \citep{gregor2014,gregor2015draw,rezende2014stochastic,kingma2014stochastic}, and are a default tool for the understanding of many physical and chemical systems. 
Despite these successes and ongoing advances, there are a number of disadvantages of variational methods that limit their power and hamper their wider adoption as a default method for statistical inference. It is one of these limitations, the choice of posterior approximation, that we address in this paper.

Variational inference requires that intractable posterior distributions be approximated by a class of known probability distributions, over which we search for the best approximation to the true posterior. The class of approximations used is often limited, e.g., mean-field approximations, implying that no solution is ever able to resemble the true posterior distribution. This is a widely raised objection to variational methods, in that unlike other inferential methods such as MCMC, even in the asymptotic regime we are unable recover the true posterior distribution. 

There is much evidence that richer, more faithful posterior approximations do result in better performance. For example, when compared to sigmoid belief networks that make use of mean-field approximations, deep auto-regressive networks use a posterior approximation with an auto-regressive dependency structure that provides a clear improvement in performance \citep{mnih2014neural}. 
There is also a large body of evidence that describes the detrimental effect of limited posterior approximations. \citet{turner_sahani2011a} provide an exposition of two commonly experienced problems. The first is the widely-observed problem of under-estimation of the variance of the posterior distribution, which can result in poor predictions and unreliable decisions based on the chosen posterior approximation. The second is that the limited capacity of the posterior approximation can also result in biases in the MAP estimates of any model parameters (and this is the case e.g., in time-series models).

A number of proposals for rich posterior approximations have been explored, typically based on structured mean-field approximations that incorporate some basic form of dependency within the approximate posterior. Another potentially powerful alternative would be to specify the approximate posterior as a mixture model, such as those developed by  \citet{jaakkola1998improving, jordan1999introduction, gershman2012nonparametric}. But the mixture approach limits the potential scalability of variational inference since it requires evaluation of the log-likelihood and its gradients for each mixture component per parameter update, which is typically computationally expensive.

This paper presents a new approach for specifying approximate posterior distributions for variational inference. We begin by reviewing the current best practice for inference in general directed graphical models, based on amortized variational inference and efficient Monte Carlo gradient estimation, in section \ref{sect:amortVI}. We then make the following contributions:
\vspace{-4mm}
\begin{itemize}
\setlength{\itemsep}{1pt}\setlength{\parskip}{0pt} \setlength{\parsep}{0pt}
\item We propose the specification of approximate posterior distributions using normalizing flows, a tool for constructing complex distributions by transforming a probability density through a series of invertible mappings (sect. \ref{sect:normFlow}). Inference with normalizing flows provides a tighter, modified variational lower bound with additional terms that only add terms with linear time complexity (sect \ref{sect:infWithFlow}).
\item We show that normalizing flows admit infinitesimal flows that allow us to specify a class of  posterior approximations that in the asymptotic regime is able to recover the true posterior distribution, overcoming one oft-quoted limitation of variational inference.
\item We present a unified view of related approaches for improved posterior approximation as the application of special types of normalizing flows (sect \ref{sect:relatedWork}).
\item We show experimentally that the use of general normalizing flows systematically outperforms other competing approaches for posterior approximation. 
\end{itemize}
\vspace{-4mm}
\section{Amortized Variational Inference}
\label{sect:amortVI}
To perform inference it is sufficient to reason using the marginal likelihood of a probabilistic model, and requires the marginalization of any missing or latent variables in the model. This integration is typically intractable, and instead, we optimize a lower bound on the marginal likelihood. Consider a general probabilistic model with observations $\vx$,  latent variables $\vz$ over which we must integrate, and model parameters $\vtheta$. We introduce an approximate posterior distribution for the latent variables $q_{\phi} ( \vz | \vx )$ and follow the variational principle \cite{jordan1999introduction} to obtain a bound on the marginal likelihood:
\begin{align}
&\log p_{\theta}(\vx)  = \log \int p_{\theta}(\vx | \vz) p(\vz)d\vz \\
& = \log \int \frac{q_{\phi}(\vz |\vx)}{q_{\phi}(\vz | \vx)} p_{\theta}(\vx | \vz) p(\vz) d\vz \\
& \!\geq \! - \KL[q_{\phi}\!(\vz | \vx ) \| p(\vz)]\!\! +\!\! \expect{q\!}{\log p_{\theta}(\vx | \vz)}\!  =\! -\!\mathcal{F}\!(\vx), \label{eq:FE}
\end{align}
where we used Jensen's inequality to obtain the final equation, $p_{\theta}(\vx | \vz)$ is a likelihood function and $p(\vz)$ is a prior over the latent variables. We can easily extend this formulation to posterior inference over the parameters $\vtheta$, but we will focus on inference over the latent variables only.
This bound is often referred to as the negative free energy $\mathcal{F}$ or as the evidence lower bound (ELBO). It consists of two terms: the first is the KL divergence between the approximate posterior and the prior distribution (which acts as a regularizer), and the second is a reconstruction error. This bound \eqref{eq:FE} provides a unified objective function for optimization of both the parameters $\vtheta$ and $\vphi$ of the model and variational approximation, respectively. 

Current best practice in variational inference performs this optimization using mini-batches and stochastic gradient descent, which is what allows variational inference to be scaled to problems with very large data sets. There are two problems that must be addressed to successfully use the variational approach: 1) efficient computation of the derivatives of the expected log-likelihood $\nabla_\phi \expect{q_\phi(z)\!}{\log p_{\theta}(\vx | \vz)}$, and 2) choosing the richest, computationally-feasible approximate posterior distribution $q(\cdot)$. The second problem is the focus of this paper. To address the first problem, we  make use of two tools: Monte Carlo gradient estimation and inference networks, which when used together is what we refer to as \textit{amortized variational inference}.
\vspace{-2mm}
\subsection{Stochastic Backpropagation}
\label{sect:infer_optim}
The bulk of research in variational inference over the years has been on ways in which to compute the gradient of the expected log-likelihood $\nabla_\phi \expect{q_\phi(z)\!}{\log p(\vx | \vz)}$. Whereas we would have previously resorted to local variational methods \citep{bishop2006pattern}, in general we now always compute such expectations using Monte Carlo approximations (including the KL term in the bound, if it is not analytically known). This forms what has been aptly named doubly-stochastic estimation \citep{titsias2014doubly}, since we have one source of stochasticity from the mini-batch and a second from the Monte Carlo approximation of the expectation.

We focus on models with continuous latent variables, and the approach we take computes the required gradients using a non-centered reparameterization of the expectation \citep{bernardo2003non, williams1992simple}, combined with Monte Carlo approximation --- referred to as \emph{stochastic backpropagation} \citep{rezende2014stochastic}. This approach has also been referred to or as \emph{stochastic gradient variational Bayes} (SGVB) \citep{kingma2014stochastic} or as affine variational inference \citep{challis2012affine}.

Stochastic backpropagation involves two steps:
\vspace{-2mm}
\begin{itemize}
\setlength{\itemsep}{1pt}\setlength{\parskip}{0pt} \setlength{\parsep}{0pt}
\item \textbf{Reparameterization}. We reparameterize the latent variable in terms of a known base distribution and a differentiable transformation (such as a location-scale transformation or cumulative distribution function). For example, if $q_\phi(z)$ is a Gaussian distribution $\mathcal{N}(z | \mu, \sigma^2)$, with $\phi = \{\mu, \sigma^2\}$, then the location-scale transformation using the standard Normal as a base distribution allows us to reparameterize $\vz$ as:
\vspace{-1mm}
\begin{equation}
z \sim \mathcal{N}(z | \mu, \sigma^2) \Leftrightarrow z = \mu + \sigma \epsilon,  \quad \epsilon \sim \mathcal{N}(0,1) \nonumber
\vspace{-1mm}
\end{equation}
\item \textbf{Backpropagation with Monte Carlo}. We can now differentiate (backpropagation) w.r.t. the parameters $\phi$ of the variational distribution using a Monte Carlo approximation with draws from the base distribution:
\vspace{-1mm}
\begin{equation}
\nabla_\phi \expect{q_\phi(z)\!}{f_\theta(z)} \Leftrightarrow  \expect{\mathcal{N}(\epsilon | 0,1)\!}{\nabla_\phi f_\theta(\mu + \sigma \epsilon)}. \nonumber
\vspace{-1mm}
\end{equation} 
\end{itemize}  
A number of general purpose approaches based on Monte Carlo control variate (MCCV) estimators exist as an alternative to stochastic backpropagation, and allow for gradient computation with latent variables that may be continuous \textit{or} discrete \cite{williams1992simple, mnih2014neural, ranganath2013black, wingate2013automated}. An important advantage of stochastic backpropagation is that, for models with continuous latent variables, it has the lowest variance among competing estimators.
\vspace{-2mm}
\subsection{Inference Networks}
A second important practice is that the approximate posterior distribution $q_\phi(\cdot)$ is represented using a recognition model or inference network \citep{rezende2014stochastic,dayan2000,gershman2014,kingma2014stochastic}. An inference network is a model that learns an inverse map from observations to latent variables. Using an inference network, we avoid the need to compute per data point variational parameters, but can instead compute a set of global variational parameters $\vphi$ valid for inference at both training and test time. This allows us to amortize the cost of inference by generalizing between the posterior estimates for all latent variables through the parameters of the inference network. The simplest inference models that we can use are diagonal Gaussian densities, 
$q_{\phi} (\vz  | \vx) = \N(\vz | \vmu_{\phi}(\vx), \diag(\vsigma_{\phi}^2(\vx)) ), \label{eq:recogModel1}$
where the mean function $\vmu_{\phi}(\vx)$ and the standard-deviation function $\vsigma_{\phi}(\vx)$ are specified using deep neural networks.

\subsection {Deep Latent Gaussian Models}
\label{sec:DLGM}
\begin{figure}[t]
\centering
\includegraphics[width=8cm]{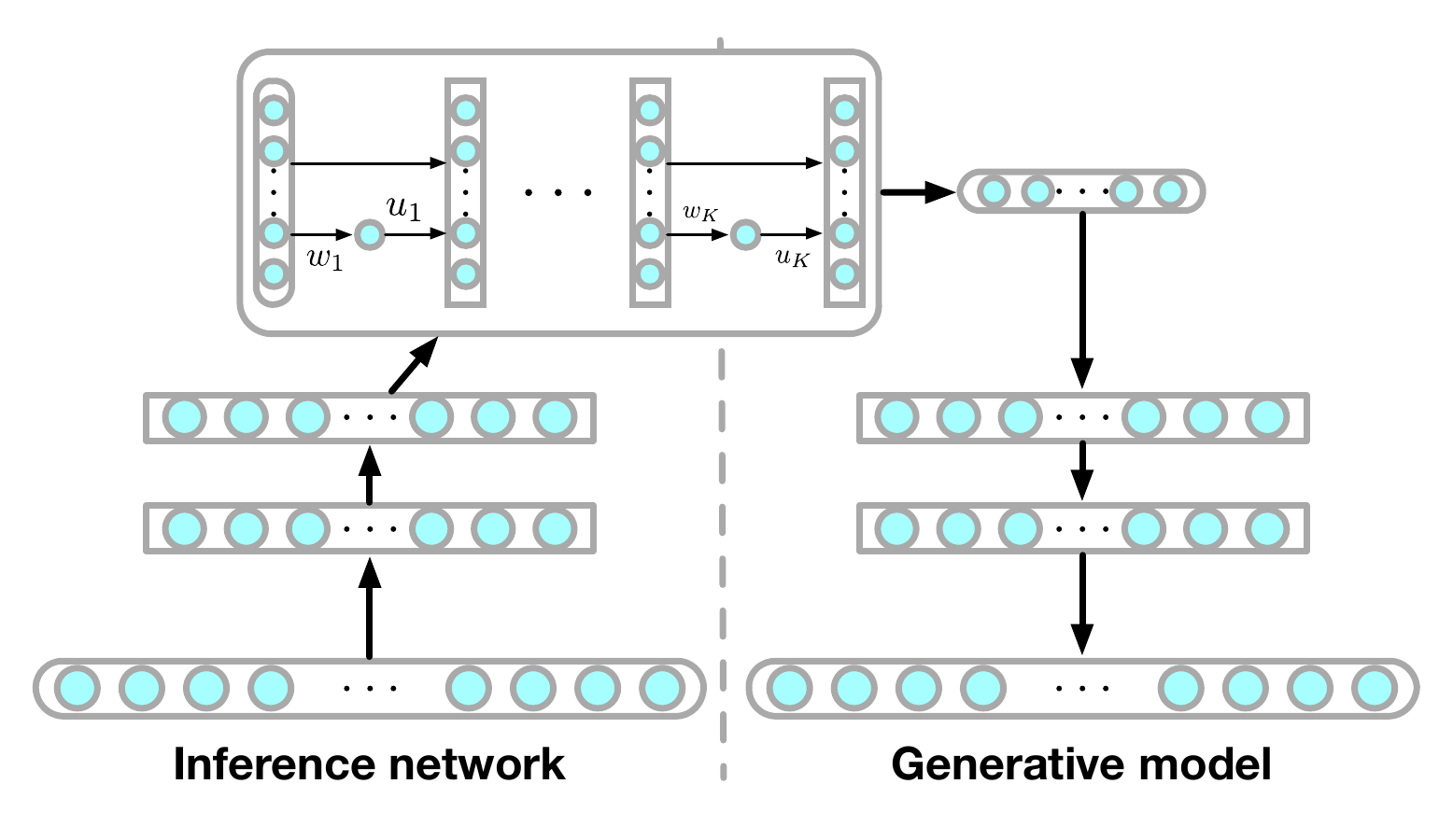}
\caption{Inference and generative models. Left: Inference network maps the observations to the parameters of the flow; Right: generative model which receives the posterior samples from the inference network during training time. 
Round containers represent layers of stochastic variables whereas square containers represent deterministic layers.}
\label{fig:arch_inference}
\end{figure}

In this paper, we study deep latent Gaussian models (DLGM), which are a general class of deep directed graphical models that consist of a hierarchy of $L$ layers of Gaussian latent variables $\vz_l$ for layer $l$. Each layer of latent variables is dependent on the layer above in a non-linear way, and for DLGMs, this non-linear dependency is specified by deep neural networks. The joint probability model is:
\begin{align}
p(\vx, \vz_1, \ldots, \vz_L) = p\left( \vx | f_0(\vz_1) \right) \prod_{l=1}^{L}p\left(\vz_{l} | f_l(\vz_{l+1})\right)
\end{align}
where the $L$th Gaussian distribution is not dependent on any other random variables. The prior over latent variables is a unit Gaussian $p(\vz_l) = \N(\vzero, \vI)$ and the observation likelihood $p_{\theta} ( \vx |\vz )$ is any appropriate distribution that is conditioned on $\vz_1$ and is also parameterized by a deep neural network (figure \ref{fig:arch_inference}). This model class is very general and includes other models such as factor analysis and PCA, non-linear factor analysis, and non-linear Gaussian belief networks as special cases \citep{rezende2014stochastic}.

DLGMs use continuous latent variables and is a model class  perfectly suited to fast amortized variational inference using the lower bound \eqref{eq:FE} and stochastic backpropagation. The end-to-end system of DLGM and inference network can be viewed as an encoder-decoder architecture, and this is the perspective taken by \citet{kingma2014stochastic} who present this combination of model and inference strategy as a variational auto-encoder. The inference networks used in \citet{kingma2014stochastic, rezende2014stochastic} are simple diagonal or diagonal-plus-low rank Gaussian distributions. The true posterior distribution will be more complex than this assumption allows for, and defining multi-modal and constrained posterior approximations in a scalable manner remains a significant open problem in variational inference.
\section{Normalizing Flows}
\label{sect:normFlow}

By examining the bound \eqref{eq:FE}, we can see that the optimal variational distribution that allows $\KL[q\|p]=0$ is one for which $q_{\phi} ( \vz | \vx )= p_{\theta} ( \vz | \vx )$, i.e. $q$ matches the true posterior distribution. This possibility is obviously not realizable given the typically used $q(\cdot)$ distributions, such as independent Gaussians or other mean-field approximations. Indeed, one limitation of the variational methodology due to the available choices of approximating families, is that even in an asymptotic regime we can not obtain the true posterior. Thus, an ideal family of variational distributions $q_{\phi} ( \vz | \vx )$ is one that is highly flexible, preferably flexible enough to contain the true posterior as one solution. One path towards this ideal is based on the principle of normalizing flows \cite{tabak2013family,tabak2010density}.

A \textit{normalizing flow} describes the transformation of a probability density through a sequence of invertible mappings. By repeatedly applying the rule for change of variables, the initial density `flows' through the sequence of invertible mappings. At the end of this sequence we obtain a valid probability distribution and hence this type of flow is referred to as a normalizing flow. 

\subsection{Finite Flows}
\label{sec:finite_flows}

The basic rule for  transformation of densities considers an invertible, smooth mapping $f:\Reals^{d} \rightarrow
\Reals^{d}$ with inverse $f^{-1} = g$, i.e. the composition  $g \circ f ( \vz ) = \vz$. If we use this mapping to transform a random variable $\vz$ with distribution $q(\vz)$, the resulting random variable $\vz'=f(\vz)$ has a distribution : 
\begin{align}
  q( \vz' ) &= q (\vz) \left| \det   \frac{\partial f^{-1}}{\partial \vz'} \right| = q (\vz) \left| \det   \frac{\partial f}{\partial \vz} \right|^{-1}, \label{eq:drule}
\end{align}
where the last equality can be seen by applying the chain rule (inverse function theorem) and is a property of Jacobians of invertible functions.
We can construct arbitrarily complex densities by composing several
simple maps and successively applying \eqref{eq:drule}. 
The density $q_{K} ( \vz )$ obtained by successively
transforming a random variable $\vz_0$ with distribution $q_{0}$ through a chain of $K$ transformations $f_{k}$ is:
\begin{align}
 \vz_K &= f_{K} \circ \ldots \circ f_2 \circ  f_{1} ( \vz_0)  \label{eq:fnComposition} \\
  \ln q_{K} (\vz_K) &=
  \ln  q_{0} ( \vz_0 ) -\sum_{k=1}^{K} \ln \left| \det  \frac{\partial f_k}{\partial \vz_{k-1} }  \right|, \label{eq:nested}
\end{align}
where equation \eqref{eq:fnComposition} will be used throughout the paper as a shorthand for the composition $f_K( f_{K-1} (\ldots f_1(x)) )$.
The path traversed by the random variables $\vz_k = f_k(\vz_{k-1})$ with initial distribution $q_0(\vz_0)$ is called the \textit{flow} and the path formed by the successive distributions $q_k$ is a \textit{normalizing flow}.
A property of such transformations, often referred to as the law of the unconscious statistician (LOTUS), is that expectations \mbox{w.r.t.} the transformed density $q_K$ can be computed without explicitly knowing $q_K$.
Any expectation $\E_{q_K}[h(\vz)]$ can be written as an expectation under $q_{0}$ as:
\begin{align}
\E_{q_K}[h(\vz)] &= \E_{ q_0 }[ h( f_{K} \circ f_{K-1} \circ   \ldots   \circ  f_{1} ( \vz_0 ) ) ],\label{eq:expectations}
\end{align}
which does not require computation of the the logdet-Jacobian terms when $h(\vz)$ does not depend on $q_{K}$.

We can understand the effect of invertible flows as  a sequence of  expansions or contractions on the initial density.
For an expansion, the map $\vz'=f(\vz)$ pulls the points $\vz$ away from a region in $\Reals^{d}$, reducing the density in that region while increasing the density outside the region. Conversely, for a contraction, the map pushes points towards the interior of a region, increasing the density in its interior while reducing the density outside.

The formalism of normalizing flows now gives us a systematic way of specifying the approximate posterior distributions $q(\vz | \vx)$ required for variational inference. With an appropriate choice of transformations $f_K$, we can initially use simple factorized distributions such as an independent Gaussian, and apply normalizing flows of different lengths to obtain increasingly complex and multi-modal distributions. 

\subsection{Infinitesimal Flows}
\label{sec:inf_flows}
It is natural to consider the case in which the length of the normalizing flow tends to infinity. In this case, we obtain an \textit{infinitesimal flow}, that is described not in terms of a finite sequence of transformations --- a finite flow, but as a partial differential equation describing how the initial density $q_0(\vz)$ evolves over `time':
$
\frac{\partial }{ \partial t} q_t(\vz) = \mathcal{T}_t[ q_t(\vz) ], \label{eq:flow_pde}
$
where $\mathcal{T}$ describes the continuous-time dynamics. 

\textbf{Langevin Flow.} One important family of flows is given by the Langevin stochastic differential equation (SDE):
\begin{align}
d\vz(t) &= \vF(\vz(t),t) dt + \vG(\vz(t),t) d\vxi(t), \label{eq:sde_flow}
\end{align}
where $d\vxi(t)$ is a Wiener process with $\E[\vxi_i(t)]=0$ and $\E[\vxi_i(t) \vxi_j(t')]=\delta_{i,j} \delta(t-t')$, $\vF$ is the drift vector and $\vD = \vG \vG^\top$ is the diffusion matrix. If we transform a random variable $\vz$ with initial density $q_0(\vz)$ through the Langevin flow \eqref{eq:sde_flow}, then the rules for the transformation of densities is given by the Fokker-Planck equation (or Kolmogorov equations in probability theory). The density $q_t(\vz)$ of the transformed samples at time $t$ will evolve as:
\begin{equation}
\frac{\partial }{ \partial t} q_t(\vz)\!\!= \!-\!\sum_i \!\frac{\partial }{ \partial z_i}[\! F_i(\vz,t) q_t ] 
+ \frac{1}{2}\!\sum_{i,j} \!\frac{\partial^2 }{ \partial z_i  \partial z_j}\! \left[ D_{ij}(\vz,t) q_t \right]. \nonumber
\end{equation}
In machine learning, we most often use the Langevin flow with $F(\vz,t) = - \nabla_{z} \mathcal{L}(\vz)$ and $G(\vz,t)=\sqrt{2} \delta_{i j}$, where $\mathcal{L}(\vz)$ is an unnormalised log-density of our model. 

Importantly, in this case the stationary solution for $q_t(\vz)$ is given by the Boltzmann distribution:
$q_{\infty}(\vz) \propto e^{-\mathcal{L}(\vz)}.$ That is, if we start from an initial density $q_0(\vz)$ and evolve its samples $\vz_0$ through the Langevin SDE, the resulting points $\vz_{\infty}$ will be distributed according to  $q_{\infty}(\vz) \propto e^{-\mathcal{L}(\vz)}$, i.e. the true posterior. This approach has been explored for sampling from complex densities by \citet{ welling2011bayesian,ahn2012bayesian, suykens1998line}.

\textbf{Hamiltonian Flow.}  Hamiltonian Monte Carlo can also be described in terms of a normalizing flow on an augmented space $\tilde{\vz} = (\vz, \vomega)$ with dynamics resulting from the Hamiltonian $\mathcal{H}(\vz, \vomega) = -\mathcal{L}(\vz) -\frac{1}{2} \vomega^\top \vM \vomega$; HMC is also widely used in machine learning, e.g., \citet{neal2011mcmc}. We will use the Hamiltonian flow to make a connection to the recently introduced Hamiltonian variational approach from \citet{salimans2014markov} in section \ref{sect:relatedWork}.

\section{Inference with  Normalizing Flows}
\label{sect:infWithFlow}
To allow for scalable inference using finite normalizing flows, we must specify a class of invertible transformations that can be used and an efficient mechanism for computing the determinant of the Jacobian. While it is straightforward to build invertible parametric functions for use in equation \eqref{eq:drule},  e.g., invertible neural networks \cite{baird2005one, rippel2013high}, such approaches typically have a complexity for computing the Jacobian determinant that scales as $O(LD^3)$, where $D$ is the dimension of the hidden layers and $L$ is the number of hidden layers used. Furthermore, computing the gradients of the Jacobian determinant involves several additional operations that are also $O(LD^3)$ and  involve matrix inverses that can be numerically unstable. We therefore require normalizing flows that allow for low-cost computation of the determinant, or where the Jacobian is not needed at all.  

\subsection{Invertible Linear-time Transformations}
We consider a family of transformations of the form: 
\begin{align}
  f ( \vz ) &=\vz+\vu h ( \vw^{\top} \vz+b ),  \label{eq:map}
\end{align}
where $\lambda = \{\vw\in \Reals^{D} , \vu \in \Reals^{D}  , b \in \Reals\}$ are free parameters  and $h ( \cdot )$ is a smooth element-wise non-linearity, with derivative $h'(\cdot)$.
For this mapping we can compute the logdet-Jacobian term in $O(D)$ time (using the matrix determinant lemma):
\begin{eqnarray}
& \psi ( \vz ) = h' ( \vw^{\top} \vz+b ) \vw \\ 
&  \left| \det    \frac{\partial f}{\partial \vz} \right| = | \det (
  \vI+\vu \psi ( \vz )^{\top} ) | = | 1+\vu^{\top} \psi ( \vz ) |.
\end{eqnarray}
From \eqref{eq:nested} we conclude that the density $q_{K} ( \vz )$ obtained by transforming an arbitrary initial density $q_0(\vz)$ through the sequence of maps $f_{k}$ of the form \eqref{eq:map} 
is implicitly given by:
\begin{align}
\vz_K & = f_{K} \circ f_{K-1} \circ   \ldots   \circ  f_{1} ( \vz ) \nonumber \\
  \ln  q_{K} (\vz_K ) &=
  \ln  q_{0} ( \vz ) - \sum_{k=1}^{K} \ln | 1+\vu_{k}^{\top} \psi_{k} ( \vz_{k-1} ) |. \label{eq:nested_maps}
\end{align}
 The flow defined by the transformation \eqref{eq:nested_maps} modifies the initial density $q_0$ by applying a series of contractions and expansions in the direction perpendicular to the hyperplane $\vw^\top\vz+b=0$, hence we refer to these maps as planar flows.

\begin{figure}[t]
\centering
\includegraphics[width=\columnwidth]{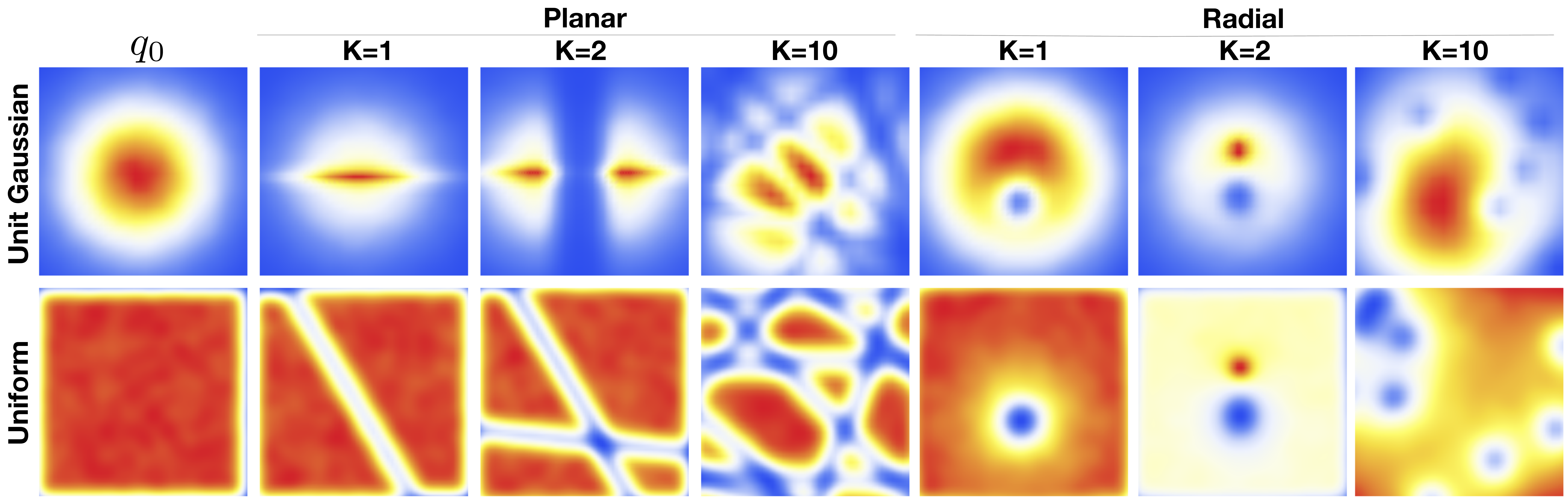}
\vspace{-4mm}
\caption{Effect of normalizing flow on two distributions.}
\label{fig:flow_example}
\vspace{-2mm}
\end{figure}

As an alternative, we can consider a family  of transformations that modify an initial density $q_0$ around a reference point $\vz_0$.
The transformation family is:
\begin{align}
& f(\vz) = \vz + \beta h( \alpha,  r) ( \vz - \vz_0 ), \label{eq:map_radial}\\
& \!\!\! \left| \det    \frac{\partial f}{\partial \vz} \right| = \left[1 + \beta h( \alpha,  r ) \right]^{d-1}  \left [ 1 + \beta h( \alpha,  r ) +   \beta h'( \alpha,  r ) r) \right], \nonumber
\end{align}
where $r=|\vz - \vz_0|$, $h(\alpha, r) = 1/(\alpha + r)$, and the parameters of the map are $\lambda=\{\vz_0 \in \Reals^D, \alpha \in \Reals^+, \beta \in \Reals\}$. This family also allows for linear-time computation of the determinant. It applies radial contractions and expansions around the reference point and are thus referred to as radial flows. We show the effect of expansions and contractions on a uniform and Gaussian initial density using the flows \eqref{eq:map} and \eqref{eq:map_radial} in figure \ref{fig:flow_example}. This visualization shows that we can transform a spherical Gaussian distribution into a bimodal distribution by applying two successive transformations.

Not all functions of the form \eqref{eq:map}  or \eqref{eq:map_radial} will be invertible. We discuss the conditions for invertibility and how to satisfy them in a numerically stable way in the appendix.

\subsection{Flow-Based Free Energy Bound}
\label{sec:rep_bound}

If we parameterize the approximate posterior distribution with a flow of length $K$, $q_{\phi}( \vz | \vx ) := q_{K} ( \vz_{K} )$, the free energy \eqref{eq:FE} can be written as an expectation over the initial distribution $q_0(\vz)$:
\begin{align}
  \FE(\vx) & = \E_{q_{\phi} ( z | x )} [ \log  q_{\phi} ( \vz | \vx ) - \log  p ( \vx, \vz ) ] \nonumber \\
  &= \E_{q_{0} ( z_0 )} \left [ \ln  q_{K} ( \vz_K ) - \log  p ( \vx ,\vz_K ) \right ] \nonumber \\
  &= \E_{q_{0} ( z_0 )} \left[ \ln  q_{0} ( \vz_0 ) \right] - \E_{q_{0} ( z_0 )} \left[\log  p ( \vx ,\vz_K ) \right] \nonumber\\
  & - \E_{q_{0} ( z_0 )} \left[ \sum_{k=1}^{K} \ln |1+\vu_{k}^{\top} \psi_{k} ( \vz_{k-1} ) | \right]. \label{eq:tbound}
\end{align}
Normalizing flows and this free energy bound can be used with any variational optimization scheme, including generalized variational EM.
For amortized variational inference, we construct an inference model using a deep neural network to build a mapping from the observations $\vx$ to the parameters of the initial density $q_0=\N(\mu, \sigma)$ ($\mu \in \Reals^D$ and $\sigma \in \Reals^D$) as well as the parameters of the flow $\lambda$. 

\subsection{Algorithm Summary and Complexity}

The resulting algorithm is a simple modification  of the amortized inference algorithm for DLGMs described by \cite{kingma2014stochastic, rezende2014stochastic}, which we summarize in algorithm \ref{alg:VAENF}. 
By using an inference network we are able to form a single computational graph which allows for easy computation of all the gradients of the parameters of the inference network and the generative model. The estimated gradients are used in conjunction with preconditioned stochastic gradient-based optimization methods such as RMSprop or AdaGrad~\citep{duchi2010adaptive}, where we use parameter updates of the form: $(\vtheta^{t+1}, \vphi^{t+1}) \gets (\vtheta^{t}, \vphi^{t}) + \boldsymbol{\Gamma}^t (\vg^t_{\theta}, \vg^t_{\phi} ), \nonumber$
with $\vGamma$ is a diagonal preconditioning matrix that adaptively scales the gradients for faster minimization.

The algorithmic complexity of jointly sampling and computing the log-det-Jacobian terms of the inference model scales as $O(L N^2) + O(KD)$, where $L$ is the number of deterministic layers used to map the data to the parameters of the flow, $N$ is the average hidden layer size, $K$ is the flow-length and $D$ is the dimension of the latent variables.
Thus the overall algorithm is at most quadratic making the overall approach competitive with other large-scale systems used in practice.
\begin{algorithm}[t]
\caption{Variational Inf. with Normalizing Flows}
\begin{algorithmic}
\STATE{Parameters: $\vphi $ variational, $\vtheta$ generative}
\WHILE{not converged}
    \STATE{$\vx \leftarrow $} \COMMENT{Get mini-batch}
    \STATE{$\vz_0 \sim q_0(\bullet | \vx)$}
    \STATE{$\vz_K \leftarrow f_{K} \circ f_{K-1} \circ   \ldots   \circ  f_{1} ( \vz_0 )$}
    \STATE{$\FE(\vx) \approx \FE(\vx,\vz_K)$}
	\STATE{$\Delta \vtheta \propto -\nabla_{\theta} \FE(\vx)$}
	\STATE{$\Delta \vphi \propto -\nabla_{\phi} \FE(\vx)$}
\ENDWHILE
\end{algorithmic}
\label{alg:VAENF}
\end{algorithm}
\section{Alternative Flow-based Posteriors}
\label{sect:relatedWork}
Using the framework of normalizing flows, we can provide a unified view of recent proposals for designing more flexible posterior approximations. At the outset, we distinguish between two types of flow mechanisms that differ in how the Jacobian is handled. The work in this paper considers \textit{general normalizing flows} and presents a method for linear-time computation of the Jacobian.  In contrast, \textit{volume-preserving flows} design the flow such that its  Jacobian-determinant is equal to one while still allowing for rich posterior distributions. Both these categories allow for flows that may be finite or infinitesimal.

The Non-linear Independent Components Estimation (NICE) developed by \citet{dinh2014nice} is an instance of a \textit{finite volume-preserving flow}. The transformations used are neural networks $f(\cdot)$ with easy to compute inverses $g(\cdot)$ of the form: 
\begin{align}
  f ( \vz ) &=( \vz_A, \vz_B + h_{\lambda} ( \vz_A ) ),  \label{eq:nice_map} \\
  g(\vz') &= ( \vz'_A, \vz'_B - h_{\lambda} ( \vz'_A ) ). \label{eq:nice_map_inverse}
\end{align}
where $\vz = (\vz_A, \vz_B)$ is an arbitrary partitioning of the vector $\vz$ and $h_{\lambda}$ is a neural network with parameters $\lambda$. 
This form results in a Jacobian that has a zero upper triangular part, resulting in a determinant of 1. 
In order to build a transformation capable of mixing all components of the initial random variable $\vz_{0}$, such flows must alternate between different partitionings of $\vz_k$. 
The resulting density using the forward and inverse transformations is given by : 
\begin{eqnarray}
  &\ln  q_{K} ( f_{K} \circ f_{K-1} \circ   \ldots   \circ  f_{1} ( \vz_0 ) ) = \ln  q_{0} ( \vz_0 ), \label{eq:nested_maps_nice} \\
  & \ln  q_{K} (\vz') = q_0( g_{1} \circ g_{2} \circ   \ldots   \circ  g_{K} ( \vz' ) ). \label{eq:nested_maps_nice_inverse}
\end{eqnarray}
We will compare NICE to the general transformation approach described in section \ref{sect:infer_optim}. \citet{dinh2014nice} assume the partitioning is of the form $\vz = [\vz_A = \vz_{1:d}, \vz_B = \vz_{d+1:D}]$. To enhance mixing of the components in the flow, we introduce two mechanisms for mixing the components of $\vz$ before separating them in the disjoint subgroups $\vz_A$ and $\vz_B$. The first mechanism applies a random permutation (NICE-perm) and the second applies a random orthogonal transformation (NICE-orth)\footnote{ Random orthogonal transformations can be generated by sampling a matrix with independent unit-Gaussian entries $A_{i,j} \sim \mathcal{N}(\vzero,\vI)$ and then performing a QR-factorization. The resulting $Q$-matrix will be a random orthogonal matrix \cite{genz1998methods}.}. 

The Hamiltonian variational approximation (HVI) developed by  \citet {salimans2014markov} is an instance of an \textit{infinitesimal volume-preserving flow}. For HVI, we consider posterior approximations $q(\vz, \vomega | \vx)$ that make use of additional auxiliary variables $\vomega$. The latent variables $\vz$ are independent of the auxiliary variables  $\vomega$ and using the change of variables rule, the resulting distribution is: $q(\vz', \vomega') = |\vJ| q(\vz)q(\vomega)$,
where $\vz', \vomega' = f(\vz, \vomega)$ using a transformation $f$. \citet {salimans2014markov} obtain a volume-preserving invertible transformation by exploiting the use of such transition operators in the MCMC literature, in particular the methods  of Langevin and Hybrid Monte Carlo. This is an extremely elegant approach, since we now know that as the number of iterations of the transition function tends to infinity, the distribution $q(\vz')$ will tend to the true distribution $p(\vz | \vx)$. This is an alternative way to make use of the Hamiltonian infinitesimal flow described in section \ref{sec:inf_flows}. 
A disadvantage of using the Langevin or Hamiltonian flow is that they require one or more evaluations of the likelihood and its gradients (depending in the number of leapfrog steps) per iteration during both training and test time.
\section{Results}
\label{sec:results}
Throughout this section we evaluate the effect of using normalizing flow-based posterior approximations for inference in deep latent Gaussian models (DLGMs). Training was performed by following a Monte Carlo estimate of the gradient of an annealed version of the free energy \eqref{eq:tbound_annealed}, with respect the model parameters $\vtheta$ and the variational parameters $\vphi$ using stochastic backpropoagation. The Monte Carlo estimate is computed using a single sample of the latent variables per data-point per parameter update.

A simple annealed version of the free energy is used since this was found to provide better results. The modified bound is:
\begin{align}
\vz_K &= f_{K} \circ f_{K-1}
  \circ   \ldots   \circ  f_{1} ( \vz ) \nonumber \\
  \FE^{\beta_t}(\vx)
  &= \E_{q^{0} ( \vz_0 )} \left [ \ln  p^{K} (\vz_K ) - \log  p ( \vx , \vz_K) \right ] \nonumber \\
  &=\E_{q_{0} ( \vz_0 )} \left[ \ln  q_{0} ( \vz_0 ) \right] - \beta_t \E_{q_{0} ( \vz_0 )} \left[\log  p ( \vx ,\vz_K ) \right] \nonumber \\ 
  &- \E_{q_{0} ( \vz_0 )} \left[ \sum_{k=1}^{K} \ln | 1+u_{k}^{T} \psi_{k} ( \vz_{k-1} ) | \right]   \label{eq:tbound_annealed}
\end{align}
where $\beta_t\in [0,1]$ is an inverse temperature that follows a schedule $\beta_t = \min(1, 0.01 + t/10000) $, going from 0.01 to 1 after 10000 iterations.

The deep neural networks that form the conditional probability between random variables consist of deterministic layers with 400 hidden units using the Maxout non-linearity on windows of 4 variables \cite{goodfellow2013maxout} . Briefly, the Maxout non-linearity with window-size $\Delta$ takes an input vector $\vx \in \Reals^d $ and computes: $\text{Maxout}(\vx)_k = \max_{ i \in \{ \Delta k, \Delta (k+1) \} } \vx_i $ for $k=0 \ldots d/ \Delta$.

We use mini-batches of 100 data points and RMSprop optimization (with learning rate $=1\times 10^{-5}$ and \mbox{momentum $=0.9$}) \cite{kingma2014stochastic, rezende2014stochastic}. Results were collected after $500,000$ parameter updates.
Each experiment was repeated 100 times with different random seeds and we report the averaged scores and standard errors.
The true marginal likelihood is estimated by importance sampling using 200 samples from the inference network as in \citep[App. E]{rezende2014stochastic}.

\subsection{Representative Power of Normalizing Flows}
\label{sec:energy}
To provide an insight into the representative power of density approximations based on normalizing flows, we parameterize a set of unnormalized 2D densities $p(\vz) \propto \exp [-U(\vz)]$ which are listed in table \ref{tab:energies}. 

\begin{table}[tbp]
\centering
\small
\caption{Test energy functions.}
\label{tab:energies}
\begin{tabular}{l}
\hline
\textbf{Potential} $U( \vz )$\\
\hline \hline
\textbf{1:}  $\frac{1}{2} \left( \frac{ \| \vz \| -2 }{0.4} \right)^{2} - \ln  
  \left( e^{- \frac{1}{2} \left[ \frac{\vz_{1} -2}{0.6} \right]^{2}} +e^{-
  \frac{1}{2} \left[ \frac{\vz_{1} +2}{0.6} \right]^{2}} \right) $\\
\textbf{2:}  $\frac{1}{2} \left[ \frac{\vz_{2} -w_{1} ( \vz )}{0.4} \right]^{2}$\\
\textbf{3:}  $-\ln   \left( e^{- \frac{1}{2} \left[ \frac{\vz_{2} -w_{1} ( \vz )}{0.35}
  \right]^{2}} +e^{- \frac{1}{2} \left[ \frac{\vz_{2} -w_{1} ( \vz ) +w_{2} ( \vz
  )}{0.35} \right]^{2}} \right)$ \\
\textbf{4:}  $-\ln   \left( e^{- \frac{1}{2} \left[ \frac{\vz_{2} -w_{1} ( \vz )}{0.4}
  \right]^{2}} +e^{- \frac{1}{2} \left[ \frac{\vz_{2} -w_{1} ( \vz ) +w_{3} ( \vz
  )}{0.35} \right]^{2}} \right)$\\
 \hline
 with $w_{1} ( \vz ) = \sin \left( \frac{2 \pi \vz_{1}}{4} \right) $, $w_{2} ( \vz )
 =3e^{- \frac{1}{2} \left[ \frac{( \vz_{1} -1 )}{0.6} \right]^{2}}$,\\ 
 $w_{3} ( \vz )=3 \sigma \left( \frac{\vz_{1} -1}{0.3} \right)$ and $\sigma ( x ) =1/ ( 1+e^{-x} )$.\\
\hline
\end{tabular}
\end{table}

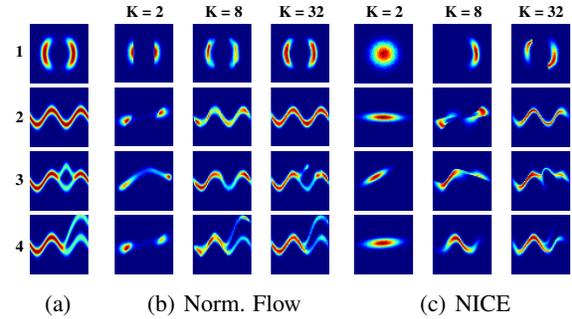
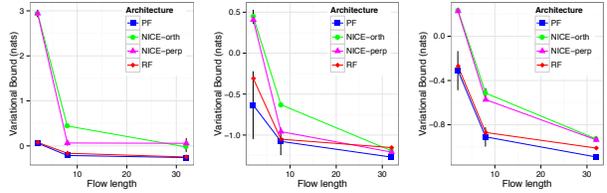
\begin{figure}[t]
\centering
\subfigure[]{%
\begin{overpic}[height=3.5cm]{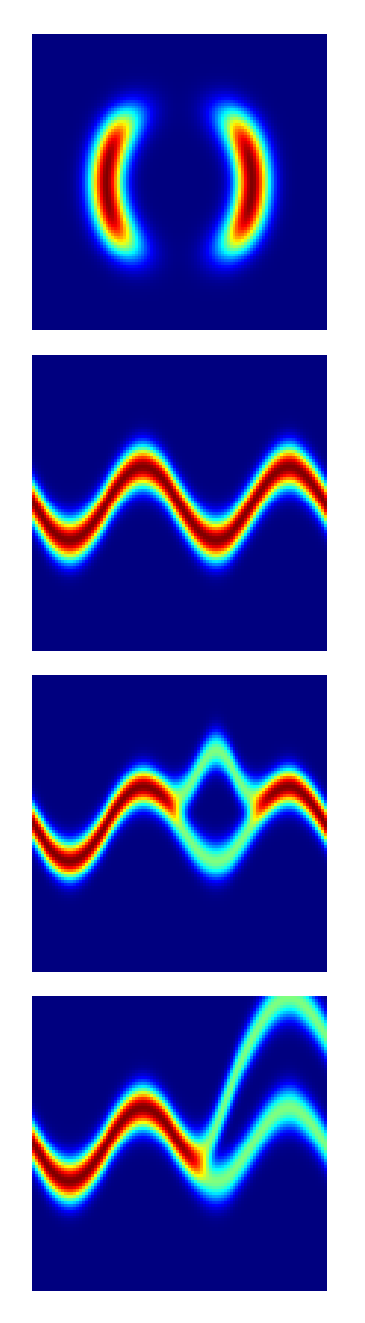} 
\put (-3,85) {\tiny \textbf{1}} \put (-3,60) {\tiny \textbf{2}} \put (-3,36) {\tiny \textbf{3}} \put (-3,11) {\tiny \textbf{4}}
\end{overpic}
\label{fig:2d_truth}}
\subfigure[Norm. Flow]{%
\begin{overpic}[height=3.5cm]{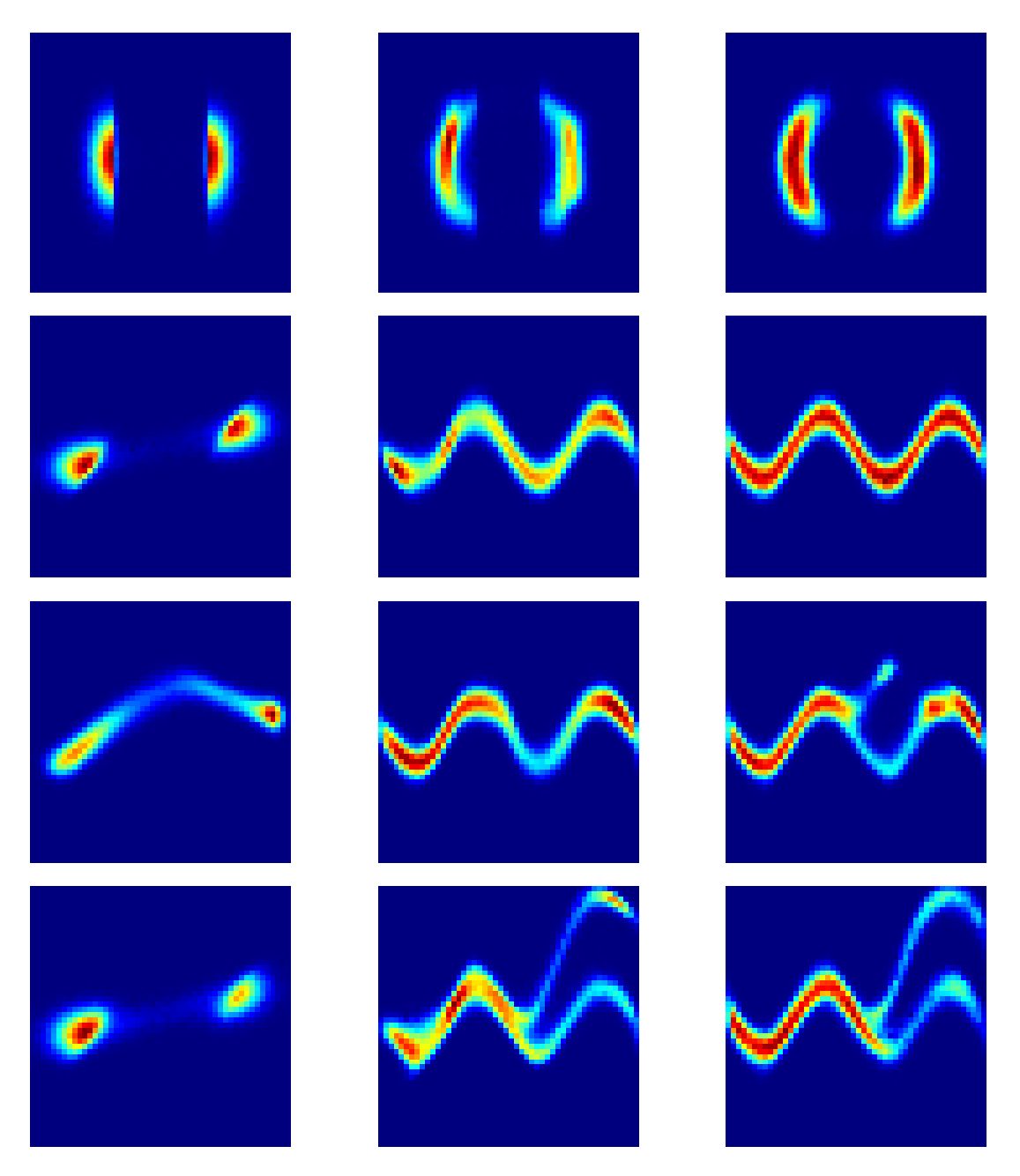} 
\put (7,100) {\tiny \textbf{K = 2}} \put (37,100) {\tiny \textbf{K = 8}} \put (65,100) {\tiny \textbf{K = 32}}
\end{overpic}
\label{fig:2D_flow_simple}}
\subfigure[NICE]{%
\begin{overpic}[height=3.5cm]{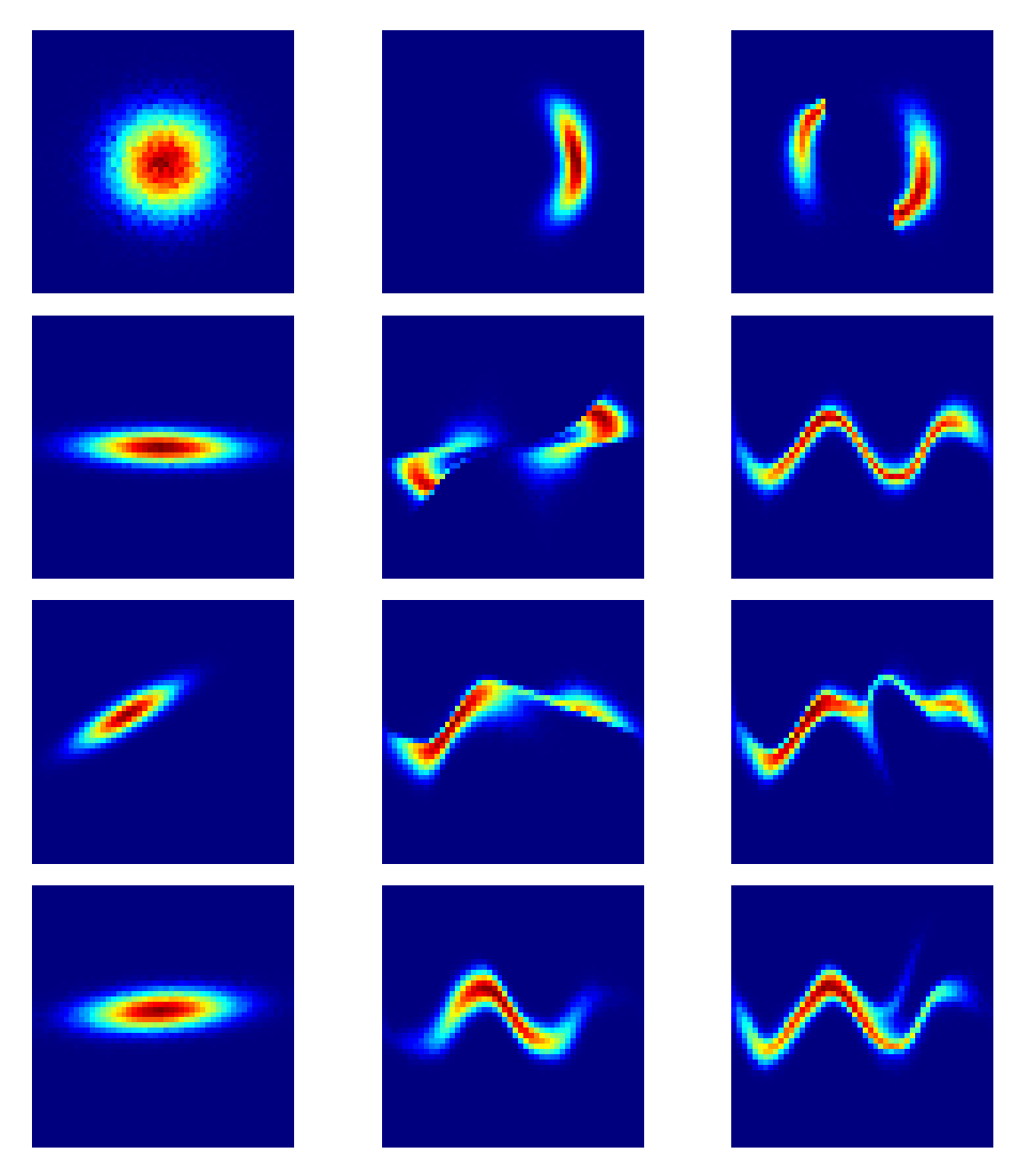} 
\put (7,100) {\tiny \textbf{K = 2}} \put (37,100) {\tiny \textbf{K = 8}} \put (65,100) {\tiny \textbf{K = 32}}
\end{overpic}

\label{fig:2d_flow_nice}}
\quad
\subfigure[Comparison of KL-divergences.]{%
\includegraphics[width = \columnwidth]{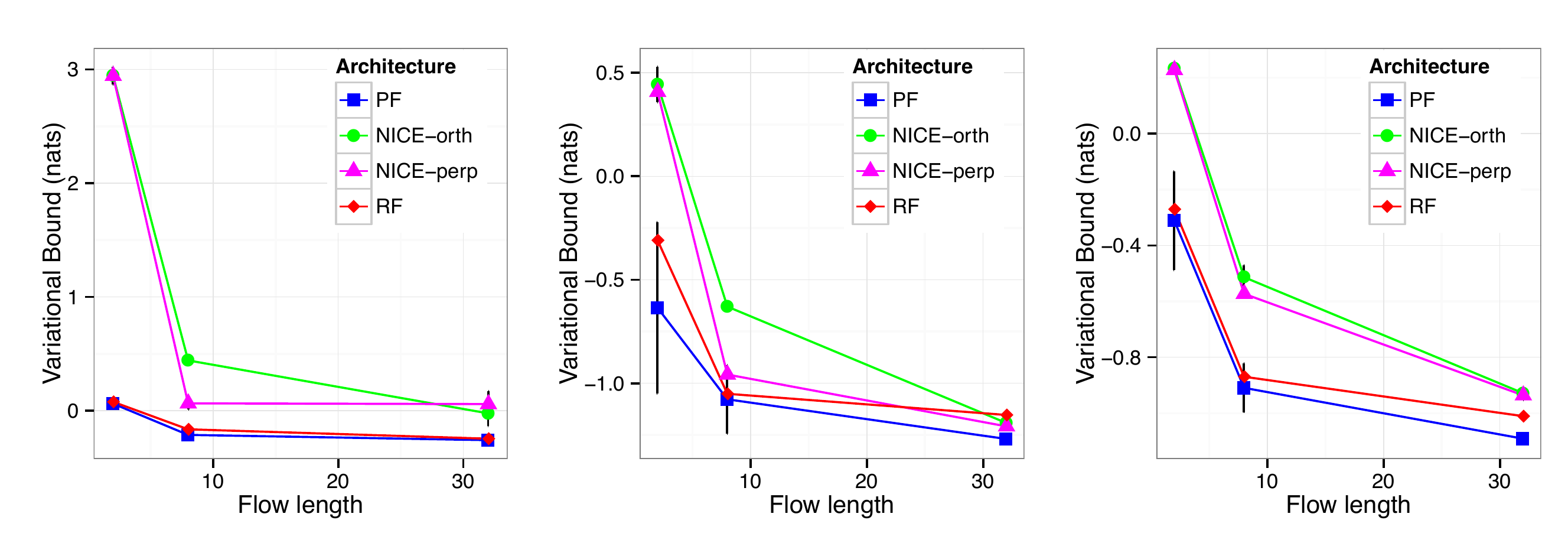}
\label{fig:2d_flow_perf}}
\caption{Approximating four non-Gaussian 2D distributions. The images represent densities for each energy function in table \ref{tab:energies} in the range $(-4,4)^2$.
(a) True posterior; (b) Approx posterior using the normalizing flow \eqref{eq:nested_maps}; (c)  Approx posterior using NICE \eqref{eq:nested_maps_nice_inverse}; (d) Summary results comparing KL-divergences between the true and approximated densities for the first 3 cases.
}
\label{fig:2D}
\vspace{-8mm}
\end{figure}

In figure \ref{fig:2d_truth} we show the true distribution for four cases, which show distributions that have characteristics such as multi-modality and periodicity that cannot be captured with typically-used posterior approximations. 

Figure \ref{fig:2D_flow_simple} shows the performance of normalizing flow approximations for these densities using flow lengths of 2, 8 and 32 transformations. The non-linearity $h(\vz) = \tanh(\vz)$ in equation \eqref{eq:map} was used for the mapping and the initial distribution was a diagonal Gaussian, $q_0(\vz) = \N(\vz|\vmu, \sigma^2 \vI)$. We see a substantial improvement in the approximation quality as we increase the flow length. Figure \ref{fig:2d_flow_nice} shows the same approximation using the volume-preserving transformation used in NICE \citep{dinh2014nice} for the same number of transformations. We show summary statistics for the planar flow \eqref{eq:nested_maps}, and NICE \eqref{eq:nested_maps_nice} for random orthogonal matrices and with random permutation matrices in \ref{fig:2d_flow_perf}.
We found that NICE and the planar flow \eqref{eq:nested_maps} may achieve the same asymptotic performance as we grow the flow-length, but the planar flow \eqref{eq:nested_maps} requires far fewer parameters. Presumably because all parameters of the flow \eqref{eq:nested_maps} are learned, in contrast to NICE which requires an extra mechanism for mixing the components that is not learned but randomly initialized. 
We did not observe a substantial difference between using random orthogonal matrices or random permutation matrices in NICE.

\subsection{MNIST and CIFAR-10 Images}
\label{sec:mnist}
The MNIST digit dataset \cite{lecun1998mnist} contains 60,000 training and 10,000 test images of ten handwritten digits (0 to 9) that are $28\times28$ pixels in size. We used the binarized dataset as in \cite{uria2013}. We trained different DLGMs with 40 latent variables for $500,000$ parameter updates.

\begin{figure}[t]
\centering
\subfigure[Bound $\FE(\vx)$]{%
\includegraphics[width=0.31\columnwidth]{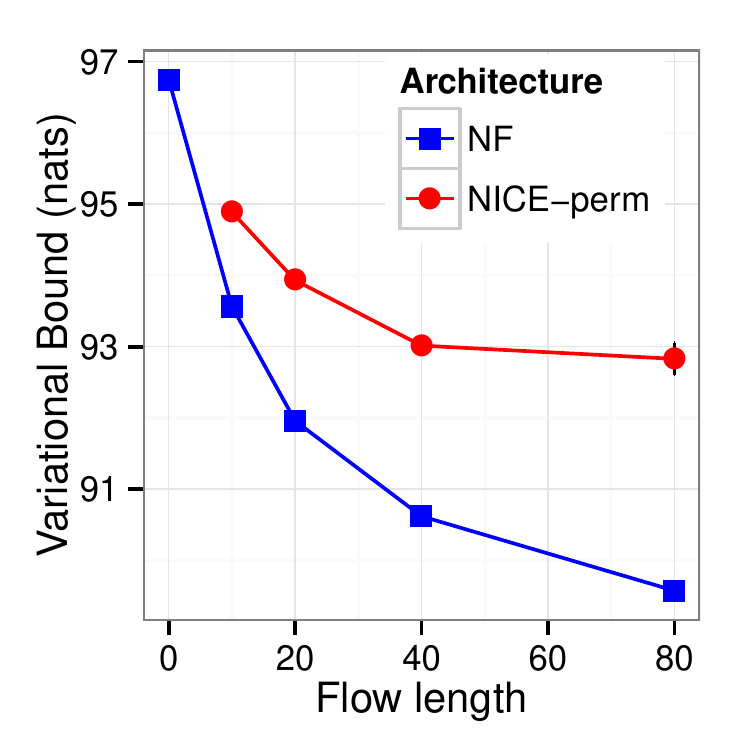}
\label{fig:bound_mnist}}
\subfigure[$\KL(q;p(z|x))$]{%
\includegraphics[width=0.31\columnwidth]{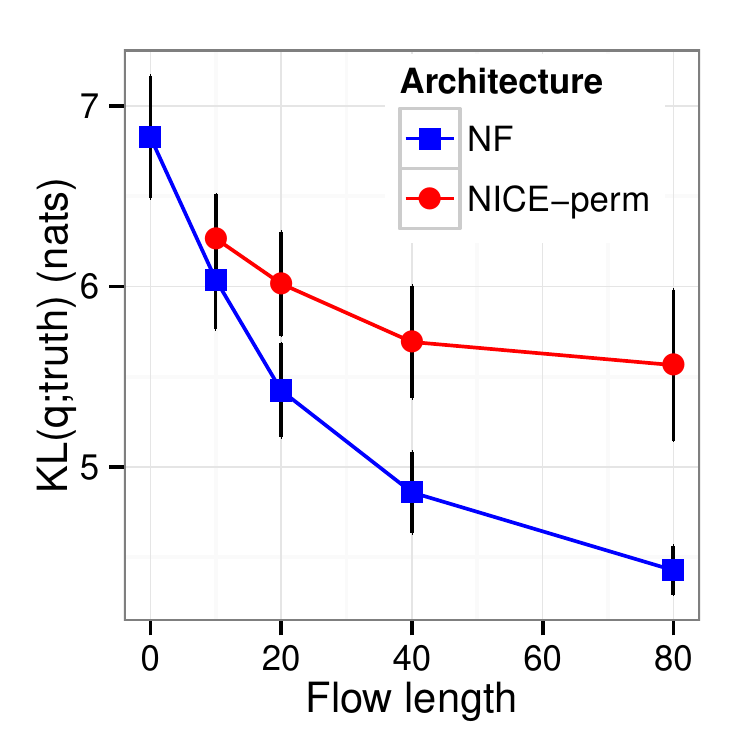}
\label{fig:kl_mnist}}
\subfigure[$- \ln p(\vx)$]{%
\includegraphics[width=0.31\columnwidth]{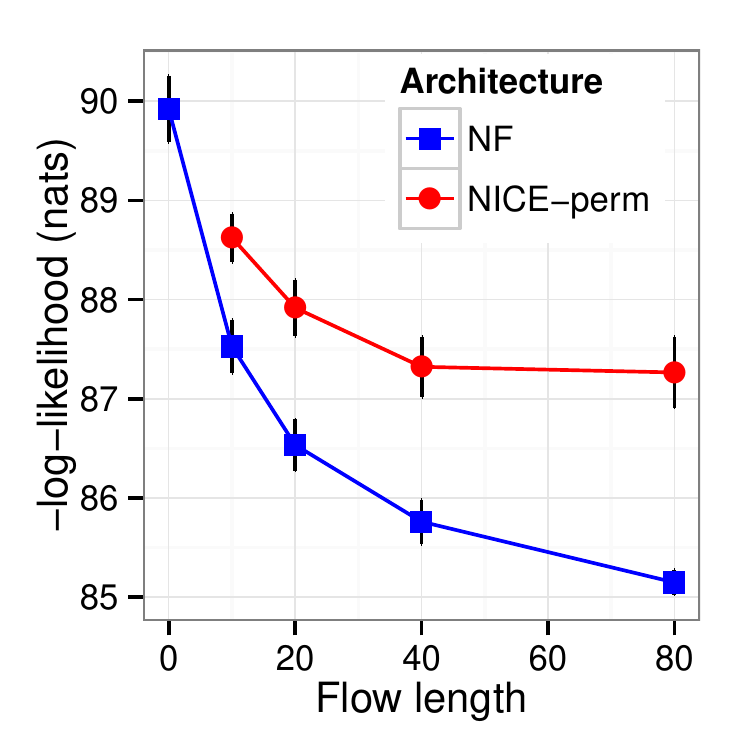}
\label{fig:ll_mnist}}
\caption{Effect of the flow-length on MNIST.}
\label{fig:perf_mnist}
\vspace{-2mm}
\end{figure}

The performance of a DLGM using the (planar) normalizing flow (DLGM+NF) approximation is compared to the volume-preserving approaches using NICE (DLGM+NICE) on exactly the same model for different flow-lengths $K$, and we summarize the performance in figure \ref{fig:perf_mnist}. This graph shows that an increase in the flow-length systematically improves the bound $\FE$, as shown in figure \ref{fig:bound_mnist}, and reduces the KL-divergence between the approximate posterior $q(\vz|\vx)$ and the true posterior distribution $p(\vz|\vx)$ (figure \ref{fig:kl_mnist}). 
It also shows that the approach using general normalizing flows outperforms that of NICE. We also show a wider comparison in table \ref{tab:mnist}. Results are included for the Hamiltonian variational approach as well, but the model specification is different and thus gives an indication of attainable performance for this approach on this data set. 

\begin{table}[tbp]
\centering
\small
\caption{Comparison of negative log-probabilities on the test set for the binarised MNIST data.}
\label{tab:mnist}
\begin{tabular}{lcc}
\hline
\textbf{Model} & & $-\ln p(\vx)$ \\
\hline \hline
DLGM diagonal covariance & & $\leq 89.9$ \\
DLGM+NF (k = 10) & & $\leq 87.5$ \\
DLGM+NF (k = 20) & & $\leq 86.5$ \\
DLGM+NF (k = 40) & & $\leq 85.7$ \\
DLGM+NF (k = 80) & & $\leq 85.1$ \\
\hline
DLGM+NICE (k = 10) & & $\leq 88.6$ \\
DLGM+NICE (k = 20) & & $\leq 87.9$ \\
DLGM+NICE (k = 40) & & $\leq 87.3$ \\
DLGM+NICE (k = 80) & & $\leq 87.2$ \\
\hline
\multicolumn{2}{c}{\tiny\textit{Results below from \citep{salimans2014markov}}} \\
DLGM + HVI (1 leapfrog step) & & $ 88.08$ \\
DLGM + HVI (4 leapfrog steps) & & $86.40$ \\
DLGM + HVI (8 leapfrog steps) & & $85.51$ \\
\hline
\multicolumn{2}{c}{\tiny\textit{Results below from \citep{gregor2014}}} \\
DARN $n_h = 500$ & &$ 84.71$\\
DARN $n_h = 500$, adaNoise  & &$ 84.13$\\
\hline
\end{tabular}
\vspace{-4mm}
\end{table}

The CIFAR-10 natural images dataset \cite{krizhevsky2010convolutional}
consists of 50,000 training and 10,000 test RGB images that are of size 3x32x32 pixels from which we extract 3x8x8 random patches. The color levels were converted to the range $[\epsilon,1-\epsilon]$ with $\epsilon=0.0001$.
Here we used similar DLGMs as used for the MNIST experiment, but with 30 latent variables. Since this data is non-binary, we use a logit-normal observation likelihood,
$
p(\vx|\vmu,\valpha) = \prod_i \frac{ \N( \text{logit}( \vx_i ) | \vmu_i, \valpha_i ) }{ \vx_i ( 1 - \vx_i ) },
$
where $\text{logit}(x)=\log \frac{x}{1-x}$. We summarize the results in table \ref{tab:cifar} where we are again able to show that an increase in the flow length $K$ systematically improves the test log-likelihoods, resulting in better posterior approximations.

\begin{table}[tbp]
\centering
\small
\caption{Test set performance on the CIFAR-10 data.}
\label{tab:cifar}
\begin{tabular}{ccccc}
\hline
 &$K=0$ & $K=2$ & $K=5$ & $K=10$\\
\hline \hline
$- \ln p(\vx)$ &-293.7& -308.6 & -317.9 &  -320.7\\
\hline
\end{tabular}
\vspace{-2mm}
\end{table}

\vspace{-2mm}
\section{Conclusion and Discussion}

In this work we developed a simple approach for learning highly non-Gaussian posterior densities by learning transformations of simple densities to more complex ones through a normalizing flow. When combined with an amortized approach for variational inference using inference networks and efficient Monte Carlo gradient estimation, we are able to show clear improvements over simple approximations on different problems. Using this view of normalizing flows, we are able to provide a unified perspective of other closely related methods for flexible posterior estimation that points to a wide spectrum of approaches for designing more powerful posterior approximations with different statistical and computational tradeoffs.

An important conclusion from the discussion in section \ref{sect:normFlow} is that there exist classes of normalizing flows that allow us to create extremely rich posterior approximations for variational inference. With normalizing flows, we are able to show that in the asymptotic regime, the space of solutions is rich enough to contain the true posterior distribution. If we combine this with the local convergence and consistency results for maximum likelihood parameter estimation in certain classes of latent variables models \citep{wang2004convergence}, we see that we are now able overcome the objections to using variational inference as a competitive and default approach for statistical inference. Making such statements rigorous is an important line of future research.

Normalizing flows allow us to control the complexity of the posterior at run-time by simply increasing the flow length of the sequence. The approach we presented considered normalizing flows based on simple transformations of the form \eqref{eq:map} and \eqref{eq:map_radial}. These are just two of the many maps that can be used, and alternative transforms can be designed for posterior approximations that may require other constraints, e.g., a restricted support. An important avenue of future research lies in describing the classes of transformations that allow for different characteristics of the posterior and that still allow for efficient, linear-time computation. 

\textbf{Ackowledgements:} We thank Charles Blundell, Theophane Weber and Daan Wierstra for helpful discussions.

\clearpage
\balance
\bibliography{bibliography}
\bibliographystyle{icml2015}

\clearpage
\nobalance
\appendix
\newpage
\section{Invertibility conditions \label{sec:inv_conditions} }

We describe the constraints required to have invertible maps for the planar and radial normalizing flows described in section 3.

\subsection{Planar flows}

Functions of the form (10)
are not always invertible depending on the non-linearity and parameters chosen. When using $h(x) = \text{tanh}(x)$, a sufficient condition for $f(\vz)$ to be invertible is that $\vw^\top \vu \geq -1$. 

This can be seen by splitting $\vz$ as a sum of a vector $\vz_{\perp}$ perpendicular to $\vw$ and a vector $\vz_{\parallel}$, parallel to $\vw$. Substituting $\vz = \vz_{\perp} + \vz_{\parallel}$ into 
 (10)
gives
\begin{align}
 f ( \vz ) &=\vz_{\perp} + \vz_{\parallel}+\vu h ( \vw^{\top} \vz_{\parallel}+b ) \label{eq:perp_parallel}.
\end{align}
This equation can be solved for $\vz_{\perp}$ given $\vz_{\parallel}$ and $\vy=f(\vz)$, having a unique solution
\begin{align}
\vz_{\perp}  &= y - \vz_{\parallel}-\vu h ( \vw^{\top} \vz_{\parallel}+b ).
\end{align} 
The parallel component can be further expanded as $\vz_{\parallel} = \alpha \frac{ \vw }{ || \vw ||^2 }$, where $\alpha \in \Reals$. The equation that must be solved for $\alpha$ is derived by taking the dot product of \eqref{eq:perp_parallel} with $\vw$, yielding the scalar equation
\begin{align}
\vw^T f ( \vz ) &=\alpha+\vw^T\vu h ( \alpha+b ) \label{eq:perp_parallel_alpha}.
\end{align}
A sufficient condition for \eqref{eq:perp_parallel_alpha} to be invertible w.r.t $\alpha$ is that its r.h.s $\alpha+\vw^T\vu h ( \alpha+b )$ to be a non-decreasing function. This corresponds to the condition $1+\vw^T\vu h' ( \alpha+b ) \geq 0 \equiv \vw^T\vu \geq -\frac{1}{h' ( \alpha+b )}$. Since $0 \leq h' ( \alpha+b ) \leq 1$, it suffices to have $\vw^T\vu \geq -1$.

We enforce this constraint by taking an arbitrary vector $\vu$ and modifying its component parallel to $\vw$, producing a new vector $\hat{\vu}$ such that $\vw^\top \hat{\vu} > -1$. 
The modified vector can be compactly written as $\hat{\vu}(\vw, \vu) = \vu + \left [ m(\vw^\top\vu) - (\vw^\top\vu) \right] \frac{\vw}{|| \vw ||^2}$, where the scalar function $m(x)$ is given by $m(x) = -1 + \log (1 + e^x)$.

\subsection{Radial flows}

Functions of the form (14)
are not always invertible depending on the values of $\alpha$ and $\beta$.
This can be seen by splitting the vector $\vz$ as $\vz = \vz_0 + r \hat{\vz}$, where $r=|\vz - \vz_0|$. Replacing this into (14) 
gives
\begin{align}
f(\vz) &= \vz_0 + r \hat{\vz} + \beta \frac{r \hat{\vz}}{\alpha+r}. \label{eq:map_radial_expanded}
\end{align}
This equation can be uniquely solved for $\hat{\vz}$ given $r$ and $\vy=f(\vz)$,
\begin{align}
\hat{\vz} &= \frac{\vy - \vz_0}{ r \left(1 + \frac{\beta}{\alpha+r} \right) }.
\end{align}
To obtain a scalar equation for the norm $r$, we can subtract both sides of \eqref{eq:map_radial_expanded} and take the norm of both sides. This gives
\begin{align}
|y-\vz_0| &= r\left(1 + \frac{\beta}{\alpha+r} \right). \label{eq:map_radial_scalar}
\end{align}
A sufficient condition for
\eqref{eq:map_radial_scalar} 
to be invertible is for its r.h.s. $r\left(1 + \frac{\beta}{\alpha+r} \right)$ to be a non-decreasing function, which implies $\beta \geq - \frac{ (r+\alpha)^2 }{\alpha}$. Since $r\geq 0$, it suffices to impose $\beta \geq -\alpha$. This constraint is imposed by reparametrizing $\beta$ as  $\hat{\beta} = -\alpha + m(\beta)$, where $m(x) = \log (1 + e^x)$.

\end{document}